\documentclass[10pt, conference, compsocconf]{IEEEtran}
\usepackage{cite}
\ifCLASSINFOpdf
  \usepackage[pdftex]{graphicx}
\else
\fi
\usepackage[cmex10]{amsmath}
\usepackage{amssymb}

\usepackage{algorithmic}
\usepackage{array}
\usepackage[tight,footnotesize]{subfigure}
\usepackage{url}
\usepackage{multirow}
\usepackage{kotex}
\usepackage{lipsum}
\usepackage{bm}

\begin{document}
\title{Modality Conversion of Handwritten Patterns by Cross Variational Autoencoders
\vspace{-5mm}}

\author{\IEEEauthorblockN{Taichi Sumi\IEEEauthorrefmark{1},
Brian Kenji Iwana\IEEEauthorrefmark{1},
Hideaki Hayashi\IEEEauthorrefmark{1},  
% and
Seiichi Uchida\IEEEauthorrefmark{1}}
\IEEEauthorblockA{\IEEEauthorrefmark{1}Advanced Information Technology, Kyushu University, Japan \\
\{brian, hideaki.hayashi, uchida\}@human.ait.kyushu-u.ac.jp\vspace{-3mm}}}


% \author{\IEEEauthorblockN{
% Anonymous\IEEEauthorrefmark{1}
% }
% \IEEEauthorblockA{\IEEEauthorrefmark{1}Affiliation}
% }
\maketitle
\begin{abstract}
This research attempts to construct a network that can convert online and offline handwritten characters to each other. The proposed network consists of two Variational Auto-Encoders (VAEs) with a shared latent space. The VAEs are trained to generate online and offline handwritten Latin characters simultaneously. In this way, we create a cross-modal VAE (Cross-VAE). During training, the proposed Cross-VAE is trained to minimize the reconstruction loss of the two modalities, the distribution loss of the two VAEs, and a novel third loss called the space sharing loss. This third, space sharing loss is used to encourage the modalities to share the same latent space by calculating the distance between the latent variables. Through the proposed method mutual conversion of online and offline handwritten characters is possible. In this paper, we demonstrate the performance of the Cross-VAE through qualitative and quantitative analysis. 
% In this paper, after showing the conversion result using the proposed network, we verify the effectiveness of unknown data of this network and its usefulness as a generator.
\end{abstract}

\begin{IEEEkeywords}
variational autoencoder; handwritten character recognition; modality conversion
\end{IEEEkeywords}
\vspace{-3mm}

\section{Introduction}
Handwritten characters inherently have two modalities: {\em image} and {\em temporal trajectory}. This is because a handwritten character image is comprised of a single or multiple strokes and each stroke is originally generated as a temporal trajectory along with the pen movement. This dual-modality is essential and unique to handwritten characters. Therefore, we can expect unique and more accurate recognition methods and applications by utilizing the dual-modality of handwritten characters. This expectation emphasizes the necessity of the methodologies to convert one modality to the other.  
\par
%
%In most document
%analysis and recognition research, these two modalities are treated
%independently. In general, offline character recognition and online character
%recognition tasks have been solved mostly by different methodologies.\par 
%
\par
Modality conversion from a temporal trajectory to an
image is so-called {\it inking}. For multi-stroke character
recognition, inking is a reasonable strategy to remove stroke-order
variations. In the past, many hybrid character recognition methods (e.g.,
\cite{Tanaka_1999}) have been proposed, where two recognition engines are used for
the original trajectory pattern and its ``inked'' image, respectively. In other
methods (e.g., \cite{Hamanaka}), the local direction of the temporal trajectory is
embedded into the inked image as an extra feature channel.\par
%
%There are several methodologies for utilizing or analyzing the dual-modality. 
%The most well-known methodology is 

\begin{figure}[t]
\begin{center}
%\vspace{-2cm}
\includegraphics[width=.8\columnwidth]{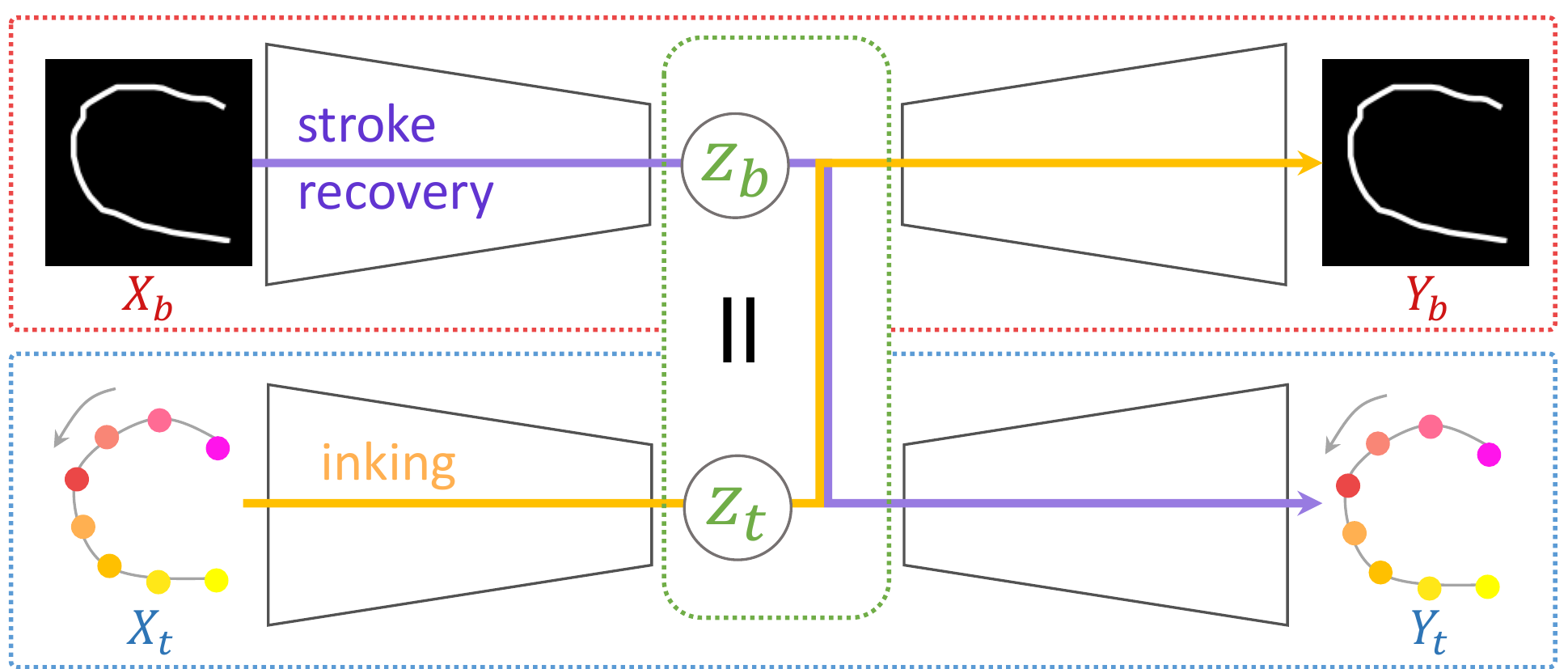}
\caption{Outline of the proposed Cross-VAE for modality conversion of handwritten characters. Two VAEs are prepared for two modalities, i.e., bitmap image and temporal trajectory, and co-trained so that their latent variables become the same for the same handwritten characters in different modalities. The trained Cross-VAE realizes inking and stroke recovery, as indicated by orange and purple paths, respectively.}
\label{fig:simpler}
\vspace{-7mm}
\end{center}
\end{figure}

Modality conversion from a handwritten character image to a
temporal trajectory representing the stroke writing order is so-called {\it stroke recovery}~\cite{Nguyen}. Comparing to the inking method, stroke
recovery is far more difficult because it is the inverse problem of inferring the
lost temporal information from a handwritten image.\par
In this paper, we propose a Cross-Variational Autoencoder (Cross-VAE), 
a neural network-based modality conversion method for handwritten
characters. Cross-VAE has the ability to convert a handwritten character image
into its original temporal trajectory and vice versa. In other words, the Cross-VAE
can realize stroke recovery as well as inking by itself. 
This means that the Cross-VAE can manage the dual-modality of handwritten characters. \par
As shown in Fig.~\ref{fig:simpler}, the Cross-VAE is compounded from two VAEs. Each VAE~\cite{vae} is a generation model
which is decomposed into two neural networks: an encoder that obtains latent 
variable $z$ from data $X$ and a decoder that obtains output $Y$ close to $X$ 
from $z$, i.e., $X\sim Y$. In general, the dimensionality of $z$ is lower than
$X$ and $Y$ and thus the latent variable $z$ represents fundamental information 
of $X$ in a compressed manner. 
One VAE of Cross-VAE is trained for a handwritten
character image (i.e., image $X_\mathrm{b} \to z_\mathrm{b} \to$ image
$Y_\mathrm{b}(\sim X_\mathrm{b})$) and the other VAE is trained for a temporal writing trajectory
(i.e., temporal trajectory $X_\mathrm{t} \to z_\mathrm{t}\to$ temporal
trajectory $Y_\mathrm{t}(\sim X_\mathrm{t})$). Note that the suffixes $\mathrm{b}$ and $\mathrm{t}$ 
indicate bitmap image and temporal trajectory, respectively.\par 
The technical highlight of Cross-VAE is that those two VAEs are trained by
considering the dual-modality of handwritten characters.  
Assume that the input image $X_\mathrm{b}$ is generated from a temporal trajectory $X_\mathrm{t}$ by inking, then we expect that their corresponding latent variables can be the same, that is, $z_\mathrm{b}=z_\mathrm{t}$. This is because $X_\mathrm{b}$ and $X_\mathrm{t}$ are the same handwritten character in different modalities and thus 
their fundamental information should be the same. Consequently, if we can co-train two VAEs under the condition $z_\mathrm{b}=z_\mathrm{t}$, we realize, for example, stroke recovery by the following steps: $X_\mathrm{b} \to z_\mathrm{b}=z_\mathrm{t} \to Y_\mathrm{t} (\sim X_\mathrm{t})$.\par

The main contributions of this paper are summarized as follows:
\begin{itemize}
    \item A cross-modal VAE is proposed for online and offline handwriting conversion. The Cross-VAE is the combination of two VAEs with different modalities with a shared latent space and a dual-modality training process. 
    \item A novel loss function called the space sharing loss is introduced. The space sharing loss encourages the latent variables of the VAEs to use the same latent space. The shared latent space is what allows for an input modality to be represented by both output modalities simultaneously.
    \item Quantitative and qualitative analyses are performed on the proposed method. We show that the Cross-VAE was able to successfully model both online and offline handwriting as well as be used for cross-modal conversion.
\end{itemize}

\section{Related Work}
 
%vae
% In recent years, there have been many proposed generation models which use neural networks and many of them involve learning latent representations of the data. 
Recently, there are two common approaches that have become popular which use neural networks to learn latent representations, Encoder-Decoders and Generative Adversarial Networks~(GAN)~\cite{gan}. 
Encoder-Decoders, such as an Autoencoder~\cite{baldi2012autoencoders}, compress data by encoding the inputs into a latent vector which is then uncompressed by the decoder. 
The Autoencoder is trained by minimizes the difference between the input and the output of the decoder.
GANs take the opposite approach and use a generator, similar to an encoder, then uses a discriminator to maximize the authenticity of the generated data. 
Where Encoder-Decoders learn the latent representations directly, GANs learn to construct data from random latent representations.

%%%%%%%%%%%%%%%%%%%%%%%%%%%%%%%%%%%%%%%%%%%
\begin{figure*}[t]
\begin{center}
\vspace{-6mm}
\includegraphics[width=0.8\textwidth]{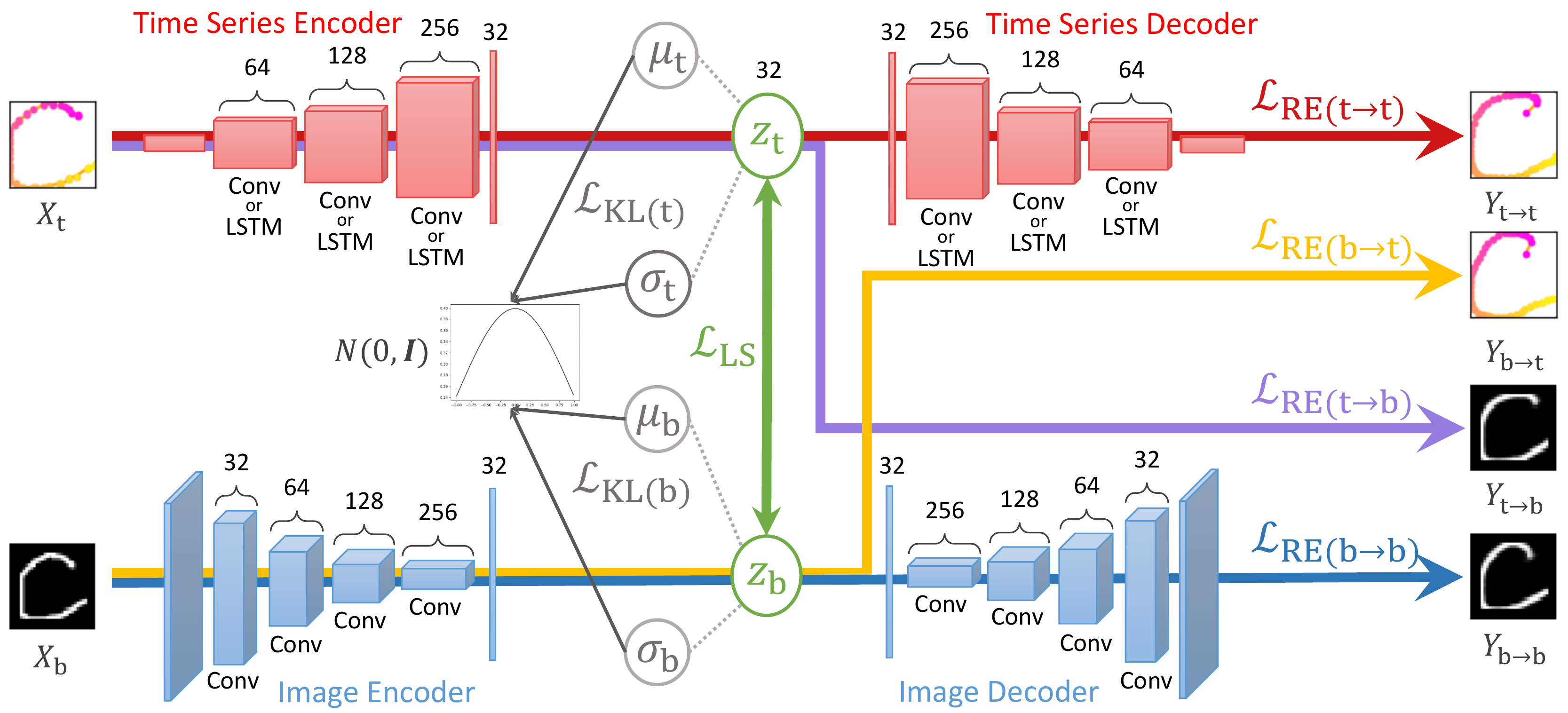}
\caption{Details of the proposed Cross-VAE. $X_\mathrm{t}$ is a time series input, $X_\mathrm{b}$ is an image input. The illustrations of the time series, $X_\mathrm{t}$, $Y_\mathrm{t \to t}$, and $X_\mathrm{b \to t}$, are colored from pink to yellow according to their sequence order.
$\mathcal{L}_\mathrm{KL}$ is the distribution loss, $\mathcal{L}_\mathrm{RE}$ is the reconstruction loss, and $\mathcal{L}_\mathrm{RE}$ is the space sharing loss. $Y_\mathrm{t \to t}$ and $Y_\mathrm{b \to b}$ are the intra-modal outputs and $Y_\mathrm{t \to b}$ and $Y_\mathrm{b \to t}$ are the cross-modal outputs.}
\label{fig:crossvae_loss}
\vspace{-4mm}
\end{center}
\end{figure*}
%%%%%%%%%%%%%%%%%%%%%%%%%%%%%%%%%%%%

% pixelCNN \cite{pixelcnn} puts an autoregressive model into the generation model that the data depends on the data generated so far, for example, by sequentially maximizing the log-likelihood of each pixel in the image, And performs generation.

% Various derivation forms exist for VAE, for example, a method of performing semi supervised learning by adding discrimination learning to the method \cite{vrae} which generates time series data and \cite{ssvae} And so on.

%cross modal
As for cross-modal generation applications, 
X-Shaped Generative Adversarial Cross-Modal Networks~(X-GACMN)~\cite{x-gacmn} creates a shared space for text and images by crossing GANs. 
Peng et al.~\cite{Peng_2019} also use GANs for text and image entanglement, however, they use weight sharing constraints.
% As another method, a method of constructing a network that converts text from images and texts from texts to \cite{linking}, learns binary codes common to pairs of images and texts, and then performs hashing of each modality There are methods such as \cite{hashing} to realize transformation by modeling.
Furthermore, a Cross-modal VAE was used by Spurr et al.~\cite{Spurr_2018} for hand pose estimation. However, their model only permits multiple pairs of encoders and decoders to share the latent space. Our method trains the VAEs to intertwine with each other and encourages them to share the same latent space.
Multi-modal and cross-modal VAEs were also used in~\cite{Huang_2018,serban2016multi}. 
Also, image-to-image translation networks can be seen as a modal conversion. 
Some examples include CycleGAN \cite{cyclegan}, StarGAN \cite{stargan}, and UNsupervised Image-to-image Translation (UNIT)~\cite{unit} networks.

%trajectory
For offline and online handwriting conversion, it has traditionally been done using classical feature-based methods~\cite{Nguyen_2010} but there has been some recent work using neural networks. 
Bhunia et al.~\cite{Kumar_Bhunia_2018} used a CNN and RNN-based Encoder-Decoder network for handwriting trajectory recovery. 
Attempts were also made using neural networks to identify graph features~\cite{Yu_Qiao_2006} and for sequential stroke prediction using regression CNNs~\cite{Zhao_2018}.

\section{Cross-modal Variational Autoencoder (Cross-VAE)}

VAEs~\cite{vae} are Autoencoders which use a variational Bayesian approach to learn the latent representation. 
% Instead of just the reconstruction loss of a typical autoencoder, the VAE uses a negative log-likelihood with a regularizer.
% Specifically, an additional distribution loss is used to minimize the difference between the distribution of the latent representation and a normal distribution. 
VAEs have been used to generate time series data~\cite{vrae}, including speech synthesis~\cite{Akuzawa_2018} and language generation~\cite{Bowman_2016}. 
They have also been used for image data~\cite{pu2016variational} and data augmentation~\cite{ssvae,wan2017variational}.

\subsection{Variational Autoencoder (VAE)}
\label{sec:vae}
A VAE~\cite{vae} consists of an encoder and a decoder. 
Given an input $X \in \mathbb{R}^I$,
the encoder estimates the posterior distribution of a latent variable $\bm{z} \in \mathbb{R}^J$.\footnote[1]{For simplicity, we omit the notation with regard to the number of training data. In the actual calculation, all losses described below are summed over the batch size.} 
% The latent representations are embedded into a latent space learned by the VAE. 
The decoder, in turn, generates an output $Y \in \mathbb{R}^I$ based on a latent variable sampled from the estimated posterior distribution. 
The VAE is trained end-to-end using a combination of the reconstruction loss $\mathcal{L}_\mathrm{RE}$ and the distribution loss $\mathcal{L}_\mathrm{KL}$, or:
\begin{equation}
    \mathcal{L}_\mathrm{VAE}=\mathcal{L}_\mathrm{KL}+\mathcal{L}_\mathrm{RE}.
\end{equation}

The reconstruction loss $\mathcal{L}_\mathrm{RE}$ is the cross-entropy between the input and the output of the decoder. 
It is determined by:
\begin{equation}
\label{eq:rec}
% \displaystyle \mathcal{L}_\mathrm{RE} &= - \mathbb{E}_{\bm{z} \sim Q(\bm{z}|X)}\left[\log{P(X|\bm{z})}\right] \nonumber \\
\displaystyle \mathcal{L}_\mathrm{RE} = - \sum_{i=1}^{I}X_i \log{Y_i}\!+\!(1\!-\!X_i)\log{(1\!-\!Y_i)},
\end{equation}
assuming that $Y$ follows the multivariate Bernoulli distribution. 
% $Q(\bm{z}|X)$ is the output of the encoder which approximates $P(\bm{z}|X)$, $L$ is the number of latent representations sampled from $Q(\bm{z}|X)$, 
In Eq.~\eqref{eq:rec}, $X_i$ and $Y_{i}$ are the $i$-th element of $X$ and $Y$, respectively.

The difference between a traditional Autoencoder or Encoder-Decoder network is that the VAE models the latent space using a Gaussian model and uses a variational lower bound to infer the posterior distribution of a latent variable. 
This is done by including a loss between the latent variables and the unit Gaussian distribution. 
Specifically, the distribution loss $\mathcal{L}_\mathrm{KL}$ is based on the Kullback-Leibler~(KL) divergence, or:
\begin{equation}
% \displaystyle \mathcal{L}_\mathrm{KL} &= \mathcal{D}\bigl(Q(\bm{z}|X)\parallel P(\bm{z})\bigr) \\
\displaystyle \mathcal{L}_\mathrm{KL} = -\frac{1}{2} \sum_{j=1}^{J} \bigl( 1+\log{(\sigma^2_j)} - \mu^2_j-\sigma^2_j \bigr),
\label{eq:kl2}
\end{equation}
assuming that the prior distribution of the latent variable $\bm{z}$ follows the multivariate Gaussian distribution of $\mathcal{N}(\bm{0},\mathbf{I})$.
In Eq.~\eqref{eq:kl2}, and $\mu$ and $\sigma^2$ are the mean and variance of the posterior distribution of $\bm{z}$. 
% $\mathcal{D}(\cdot)$ is the KL divergence between $Q(\bm{z}|X)$ and $P(\bm{z})$, 
% $J$ is the dimensionality of the latent space, 

\subsection{Cross-VAE}

We propose the use of a Cross-modal VAE~(Cross-VAE) to be used to perform online and offline handwritten character conversion, as illustrated in Fig.~\ref{fig:crossvae_loss}. 
The network in red is a VAE for online handwritten characters and the network in blue is for offline handwritten characters.
The Cross-VAE is constructed from the joining of two different single modality VAEs into one multi-modal VAE with a shared cross-modal latent space. 
Furthermore, we use a cross-modal loss function to ensure that the latent space is shared between the modalities. 

%\subsection{Shared Latent Space}
%We construct a latent space shared between multiple domains using a proposed model with VAE modified (hereinafter referred to as Cross-VAE) and perform online / offline handwritten character conversion.
%Cross-VAE is constructed as a generation model capable of handling domains (online handwritten characters) including time information and domains (time offline handwritten characters) whose time information has been lost, among cross-modal models.
%In this paper, we construct a shared latent space of online / offline handwritten characters using Cross-VAE, which is a cross-modal model, and try to convert between domains from latent variables obtained from that space.

%In Fig.~\ref{fig:crossvae_loss}, 
%By making distributions of latent variables $z_\mathrm{t}$ and $z_\mathrm{b}$ given by each encoder closer, it is possible to construct shared latent space with two VAEs and enable mutual conversion I will try.

During training, the two modalities are trained simultaneously. 
A time series input $X_\mathrm{t}$ and an image input $X_\mathrm{b}$ are entered into the encoders and four outputs are extracted from the decoders. 
For each input $X_\mathrm{t}$ and $X_\mathrm{b}$, there are respective time series outputs, $Y_\mathrm{t \to t}$ and $Y_\mathrm{b \to t}$, and respective image outputs $Y_\mathrm{t \to b}$ and $Y_\mathrm{b \to b}$. 
The outputs $Y_\mathrm{t \to t}$ and $Y_\mathrm{b \to b}$ are intra-modal and the outputs $Y_\mathrm{t \to b}$ and $Y_\mathrm{b \to t}$ are cross-modal. 

The loss function of the Cross-VAE is:
\begin{align}
\mathcal{L}_\mathrm{Cross} = \mathcal{L}_\mathrm{KL}+\mathcal{L}_\mathrm{RE}+\mathcal{L}_\mathrm{LS},
\label{eq:all}
\end{align}
where $\mathcal{L}_\mathrm{KL}$ is the  
distribution loss and $\mathcal{L}_\mathrm{RE}$ is the reconstruction loss as described in Section~\ref{sec:vae}. 
The third loss, $\mathcal{L}_\mathrm{LS}$, is the proposed space sharing loss.
Due to training with the two inputs, $X_\mathrm{t}$ and $X_\mathrm{b}$, two latent representations are created $\bm{z}_\mathrm{t}$ and $\bm{z}_\mathrm{b}$, respectively. 
Therefore, the traditional VAE losses, $\mathcal{L}_\mathrm{KL}$ and $\mathcal{L}_\mathrm{RE}$, need to be modified for Cross-VAE. 
 
Due to the two latent representations, the total distribution loss $\mathcal{L}_\mathrm{KL}$ is calculated by combining the individual distribution losses, $\mathcal{L}_\mathrm{KL(t)}$ and $\mathcal{L}_\mathrm{KL(b)}$, or:
\begin{align}
\mathcal{L}_\mathrm{KL} = \alpha \mathcal{L}_\mathrm{KL\mathrm{(t)}} + \beta
\mathcal{L}_\mathrm{KL\mathrm{(b)}},
\label{eq:klloss}
\end{align}
where $\alpha$ and $\beta$ are weights.
% Distribution distance of on-line handwritten characters VAE 
% \begin{gather}
% \begin{split}
% \displaystyle \mathcal{L}_\mathrm{KL(ts)} &= \mathcal{D}\bigl(Q(z_\mathrm{t}|X_\mathrm{t})\parallel P(z_\mathrm{t})\bigr) \\
% \displaystyle &= -\frac{1}{2} \sum_{j=1}^{J} \bigl( 1+\log{(\sigma^2_{\mathrm{t}_j})} - \mu^2_{\mathrm{t}_j}+\sigma^2_{\mathrm{t}_j} \bigr),
% \label{eq:kl2}
% \end{split}
% \end{gather}
% assuming that probability distribution $P(z_\mathrm{t})$ of latent variable $z_\mathrm{t}$ of VAE follows the multivariate Gaussian distribution of $\mathcal{N}(0,\mathbf{I})$.
% In expression \eqref{eq:kl2}, $J$ is the number of data, $\mu_\mathrm{t}$ and $\sigma_\mathrm{t}$ are $Q(z_\mathrm{t}|X_\mathrm{t})$, which is the average and variance. 
% The same applies to $\mathcal{L}_\mathrm{KL(b)}$.
The distribution loss of the individual input modalities is calculated using Eq.~\ref{eq:kl2}.

\begin{figure}[t]
	\centering
    \subfigure[Online Handwriting]{
		\includegraphics[width=0.46\columnwidth, clip,trim={0 2.2cm 0 0}]{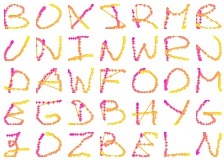} %1.1
	}
    \subfigure[Offline Handwriting]{
		\includegraphics[width=0.46\columnwidth, clip,trim={0 2.2cm 0 0}]{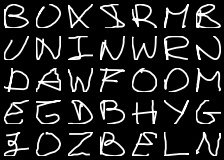} %1.1
	}

		\caption{Examples of images created from time series in the experiments. In (a), pink indicates the beginning of the sequence.}
		\label{fig:characters}
		\vspace{-3mm}
\end{figure}

\begin{figure*}[t]
    \centering
    \subfigure[Result using LSTM layers for the online encoder and decoder]{
        \includegraphics[width=1\textwidth]{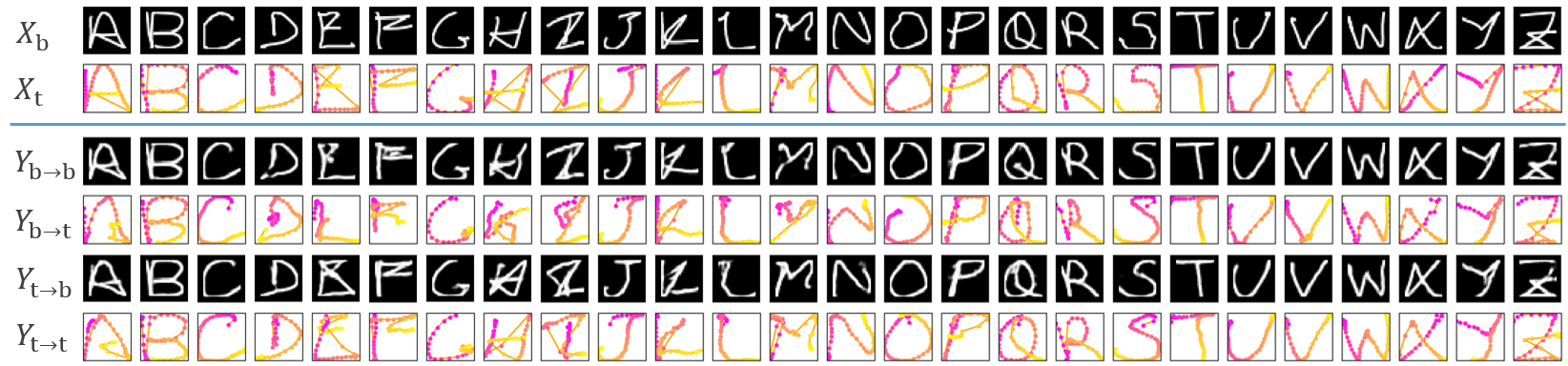}
        \label{fig:result_lstm}
        }
    \subfigure[Result using convolutional layers for the online encoder and decoder]{
        \includegraphics[width=1\textwidth]{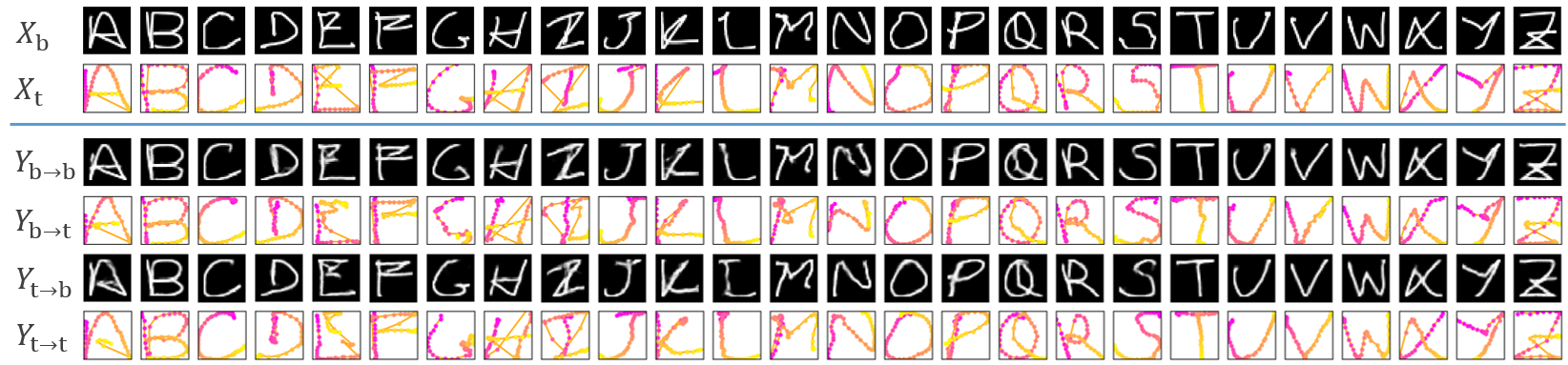}
        \label{fig:result_conv}
        }
    
\caption{Result of the Cross-VAE. $X_\mathrm{b}$ is the original image and $X_\mathrm{t}$ is the original time series. $Y_\mathrm{b \to b}$ and $Y_\mathrm{t \to t}$ are outputs of the Cross-VAE which correspond to the same modalities and $Y_\mathrm{b \to t}$ and $Y_\mathrm{t \to b}$ are between different modalities. The illustrations of the time series, $X_\mathrm{t}$, $Y_\mathrm{t \to t}$, and $X_\mathrm{b \to t}$ are colored from pink to yellow according to their sequence order.}
\label{fig:results}
\vspace{-3mm}
\end{figure*}

Next, the reconstruction loss $\mathcal{L}_\mathrm{RE}$ takes into account the reconstruction of $Y_\mathrm{t \to t}$ and $Y_\mathrm{b \to b}$, as well as the conversion of $Y_\mathrm{t \to b}$ and $Y_\mathrm{b \to t}$. 
Thus:
\begin{gather}
\begin{split}
\displaystyle \mathcal{L}_\mathrm{RE} = \gamma_\mathrm{t \to t} \mathcal{L}_\mathrm{RE\mathrm{(t \to t)}} + \gamma_\mathrm{b \to b} \mathcal{L}_\mathrm{RE\mathrm{(b \to b)}} \\
\displaystyle + \gamma_\mathrm{t \to b} \mathcal{L}_\mathrm{RE\mathrm{(t \to b)}} + \gamma_\mathrm{b \to t}\mathcal{L}_\mathrm{RE\mathrm{(b \to t)}}, \label{eq:rec_all}
\end{split}
\end{gather}
where $\mathcal{L}_\mathrm{RE\mathrm{(t \to t)}}$ and $\mathcal{L}_\mathrm{RE\mathrm{(b \to t)}}$ are the losses calculated by Eq.~\eqref{eq:rec} to  input $X_\mathrm{t}$ and $\mathcal{L}_\mathrm{RE\mathrm{(b \to b)}}$ and $\mathcal{L}_\mathrm{RE\mathrm{(t \to b)}}$ are to input $X_\mathrm{b}$. 
Also, $\gamma_\mathrm{t \to t}$, $\gamma_\mathrm{b \to b}$, $\gamma_\mathrm{t \to b}$, $\gamma_\mathrm{b \to t}$ are weight of each respective loss.
% In the expression \eqref{eq:rec_all}, the restoration error of the on-line handwritten character VAE denoted by $\mathcal{L}_\mathrm{RE(t \to t)}$ is expressed by
% \begin{equation}
% \label{eq:rec}
% \begin{split}
% \displaystyle \mathcal{L}_\mathrm{RE(t \to t)} &= - \mathrm{average}\left(\log{P(X_\mathrm{t}|z_\mathrm{t})}_{z_\mathrm{t} \sim Q_\mathrm{t}}\right) \\
% \displaystyle &\simeq - \frac{1}{J} \sum_{j=1}^{J} \sum_{i=1}^{D_\mathrm{t}}x_{\mathrm{t}_{j_i}} \log{y_{\mathrm{t}_{j_i}}}+(1-x_{\mathrm{t}_{j_i}})\log{(1-y_{\mathrm{t}_{j_i}})},
% \end{split}
% \end{equation}
% assuming that $P(X_\mathrm{t}|z_\mathrm{t})$ follows the multivariate Bernoulli distribution.
% The same applies to $\mathcal{L}_\mathrm{RE\mathrm{(b \to b)}}$, $\mathcal{L}_\mathrm{RE\mathrm{(t \to b)}}$ and $\mathcal{L}_\mathrm{RE\mathrm{(b \to t)}}$.

\subsection{Space Sharing Loss}
While the Cross-VAE is trained using the combination of the reconstruction and distribution losses for the different modalities, we propose the use of a space sharing loss function to encourage the latent variable to share the same latent space. 
The space sharing loss $\mathcal{L}_\mathrm{LS}$ gives the square error of the latent variable $\bm{z}_\mathrm{t}$ obtained from the online character VAE and the latent variable $\bm{z}_\mathrm{b}$ of the offline character VAE.
Specifically:
\begin{align}
\mathcal{L}_\mathrm{LS} = \delta \frac{1}{2} \|\bm{z}_{\mathrm{t}}-\bm{z}_{\mathrm{b}}\|^2,
\label{eq:ls}
\end{align}
where $\delta$ is a weight and $\|\cdot\|$ is the Euclidean norm. 

\section{Online and Offline Conversion of Handwritten Characters using Cross-VAE}
\subsection{Dataset}

% \begin{figure}[t]
%     \begin{subfigure}{1\columnwidth}
%     \centering
%         \includegraphics[width=1\linewidth]{figures/image_vae.png}
%         \caption{Image VAE}
%         \label{fig:img-vae}
%     \end{subfigure}
%     \begin{subfigure}{1\columnwidth}
%     \centering
%         \includegraphics[width=1\linewidth]{figures/timeseries_vae.png}
%         \caption{Time Series VAE}
%         \label{fig:ts-vae}
%     \end{subfigure}
% \caption{Details of the VAEs used in the Cross-VAE.}
% \label{fig:vaes}
% %\vspace{-2mm}
% \end{figure}

For the experiment, we used handwritten uppercase characters from the Unipen online handwritten character dataset~\cite{unipen}. 
The online handwritten characters consist of time series made of $(x,y)$ coordinates. 
The online characters were normalized to fit within a square bound by $(0,0)$ and $(1,1)$.
In order to use a second modality, the online characters were rendered into images. 
The images were $32\times32$ pixels with 0 as the background and 1 as the foreground. 
Examples of the image renderings can be found in Fig.~\ref{fig:characters}. 

%Fig.~\ref{fig:characters} shows an example of handwritten characters which are experimental data of this paper.
%Note that this is image data created from time series data as described above.
%In the original time series data, there were few data including elements called pen up pen down, and a lot of data consisted of one stroke.
%Therefore, in the experiment, as shown in %Fig.~\ref{fig:characters}, 
%outlined image data is written offline on handwritten character data on black spots where each coordinate point of time series data is plotted large, off-line handwritten character data, Time series data consisting of $x,y$ coordinates was taken as on-line handwritten character data.

\subsection{Architecture Details}
The image-based encoder and decoder were constructed from a Convolutional Neural Network~(CNN) with a similar structure as a ConvDeconv network~\cite{Noh_2015}. 
The image encoder consists of four $3\times3$ convolutional layers with Rectified Linear Unit (ReLU) activations and corresponding $2\times2$ maxpooling layers. 
The number of nodes are detailed in Fig.~\ref{fig:crossvae_loss}. 
The decoder is a reflection of the encoder which uses unpooling and deconvolutions. 
Between the convolutional layers, there exist three fully-connected layers. 
One belonging to each, the encoder and decoder, and one for the latent variable.

For the time series-based encoder and decoder, there were two architectures chosen. 
The first is a CNN-based approach with 1D convolutions and no pooling. 
The second is a Recurrent Neural Network (RNN) approach using Long Short Term Memory (LSTM)~\cite{Hochreiter_1997} layers. 
Both the CNN-based approach and the LSTM-based approach have three fully-connected layers, one for the encoder, one for the latent variable, and one for the decoder. The two layer types were chosen to compare the difference between the LSTM layers which were designed specifically for time series and convolutional layers which are traditionally used for images.

%In the experiment, it was confirmed whether mutual conversion is possible, and at the same time, which of Recurrent NN (RNN) and Convolutional NN is more suitable for this task was compared.
%Therefore, on-line handwritten character VAE was constructed using Long Short Term Memory (LSTM) layer and Convolution layer respectively.
%LSTM is an extension of RNN corresponding to time-series dependence by having self loop in intermediate layer and including regression element, and it is a network model which can be characterized as being able to deal with long-term dependency.

%Fig.~\ref{fig:vaes} shows the configuration of two VAEs included in the cross-VAE used in the experiment.
% The on-line handwritten character VAE consists of an encoder $E_\mathrm{b}$ consisting of Convolution 4 layer and Fully Connected 2 layer, and a decoder $D_\mathrm{b}$ consisting of Fully Connected 1 layer and Deconvolution 4 layer.
% The offline handwritten character VAE consists of an encoder $E_\mathrm{t}$ consisting of LSTM 3 layer or Convolution 3 layer · Fully Connected 2 layer and a decoder $D_\mathrm{t}$ consisting of Fully Connected 1 layer and LSTM 4 layer or Convolution 4 layer.
% For the entire network, the activation function is ReLU and the algorithm of the gradient descent method is RMSProp.
The Cross-VAE was optimized with RMSProp~\cite{hinton2012neural} for 200 epochs. 
The weighting factors of each loss function were determined through experiments. 
Specifically, they are $\alpha=0.5$, $\beta=0.5$, $\gamma_\mathrm{t \to t}=0.4$, $\gamma_\mathrm{b \to b}=0.5$, $\gamma_\mathrm{t \to b}=0.4$, $\gamma_\mathrm{b \to t}=0.2$, $\delta=1.0$. 
The number of dimensions of the latent variable was 32 in all experiments.

\subsection{Conversion Result}

The results of the Cross-VAE are shown in Fig.~\ref{fig:results}. 
Fig.~\ref{fig:results}~(a) is from using LSTM layers for the online encoder and decoder and Fig.~\ref{fig:results}~(b) is from using convolutional layers in the online encoder and decoder. 
The results $Y_\mathrm{b \to b}$ and $Y_\mathrm{t \to b}$ are the images generated by the inputs $X_\mathrm{t}$ and $X_\mathrm{b}$, respectively.
The results $Y_\mathrm{t \to t}$ and $Y_\mathrm{b \to t}$ are renderings of the time series colored from pink to yellow in chronological order. 
Notably, the output $Y_\mathrm{b \to t}$ is the trajectory prediction based on the image input $X_\mathrm{b}$. 

By examining Fig.~\ref{fig:results}, it can be seen that the mutual conversion of the modalities was accurately performed. 
This shows that the shared latent space learned by the simultaneous encoding of $X_\mathrm{b}$ and $X_\mathrm{t}$ is able to accurately represent both image data and time series data.
In addition, not only was the stroke trajectory inferred, the results show that the shared latent space was able to encode temporal information about what is expected from the characters. 
For example, the ``B" in Fig.~\ref{fig:results}~(a) is missing information, yet the time series results $Y_\mathrm{b \to t}$ and $Y_\mathrm{t \to t}$ were able to restore the character. 
The results from Fig.~\ref{fig:results} qualitatively confirm that the Cross-VAE is able to do mutual modality conversion between the online and offline handwritten characters.

\begin{figure}[t]
    \centering
        \includegraphics[width=1\columnwidth,clip,trim={0cm 0cm 14.3cm 0cm}]{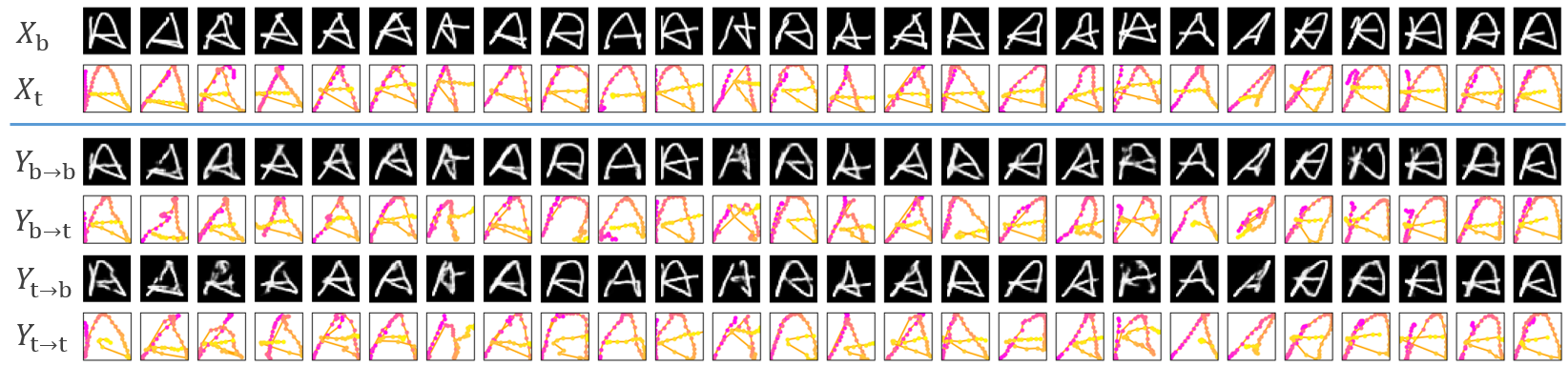}
\caption{Multiple example results for the letter ``A" using convolutional layers for the online encoder and decoder}
\label{fig:resultsextra}
\vspace{-2mm}
\end{figure}

The letter ``A" is another character that would normally be difficult to recover lost time series information due to having multiple variations. 
In some cases, the left-most stroke is drawn downwards and in some, it is drawn upwards depending on the author. 
Fig.~\ref{fig:resultsextra} are examples of many different ``A"s generated by the Cross-VAE. 
The figure shows that the Cross-VAE was able to correctly estimate most of the strokes of the ``A"s. 
In particular, the results from $Y_{b \to t}$ was able to not only correctly predict the stroke order but also was able to replicate the stroke velocity. 
Note the stroke that crosses the center of the ``A." 
This further enforces the success of the proposed Cross-VAE.

\subsection{Quantitative Evaluation of Conversion}
% Through quantitative evaluation, we consider which of the two experimental networks is suitable for the task of this paper.
% From the results in 5.2, we confirmed that mutual conversion of online and offline handwritten characters can be performed using LSTM and Convolution as online handwritten characters VAE.
% We prepared two networks for online handwritten characters VAE in order to see which model is more suitable for mutual conversion of characters.
% Therefore, quantitative evaluation is performed with each indicator for mutual conversion.

In order to evaluate the method quantitatively, we constructed the following three measures to determine the quality of the generated characters:

\paragraph{PSNR} Peak signal-to-noise ratio (PSNR) calculates the similarity between the input images and the generated output images. 
PSNR is the ratio between the maximum luminance MAX and the amount of noise, or:
\begin{equation}
\label{eq:psnr}
    \mathrm{PSNR} = 10\log_{10} \frac{\mathrm{MAX}^2}{\mathrm{MSE}},
\end{equation}
where MSE is the mean squared error between $X_{b}$ and $Y_{t \to b}$. 
PSNR is measured in decibels (dB) with a larger value being better. 
% Peak Signal-to-Noise Ratio (PSNR) and Structual Similarity (SSIM) were used as the evaluation index of inking.
% PSNR and SSIM are both indices of image evaluation, PSNR evaluates the difference in pixel value of two images, and SSIM evaluates by using pixel value, contrast, structure difference.

\paragraph{SSIM} Structural Similarity (SSIM) predicts the perceived difference between images. 
Similar to PSNR, this acts as a similarity measure between $X_{b}$ and $Y_{t \to b}$. 
The equation for SSIM is:
\begin{equation}
 \mathrm{SSIM}=
 \frac{{\left ( 2 \mu_{X_{b}}\mu_{Y_{t \to b}}+ C_1 \right )+ \left (2 \sigma_{{X_{b}}{Y_{t \to b}}}+C_2\right)}} 
{\left(\mu_{X_{b}}^2+\mu_{Y_{t \to b}}^2+C_1)(\sigma_{X_{b}}^2+\sigma_{Y_{t \to b}}^2+C_2\right)},
  \label{eq:ssim}
\end{equation}
where $C_1$ and $C_2$ are stabilizing constants set to $C_1=(0.01\times255)^2$ and $C_2=(0.03\times255)^2$. 
$\mu$ is the average luminance, $\sigma^2$ is the variance, and $\sigma$ is the covariance. 
SSIM is a value from 0 to 1 with a larger value meaning more similar. 

\paragraph{DTW} Dynamic time warping (DTW) was used as an evaluation for the time series generation as a method of measuring the stroke trajectory estimation. 
DTW is a robust distance measure between time series which uses dynamic programming to optimally match sequence elements. 
In this case, we use the average DTW-distance between the input time series $X_\mathrm{t}$ and the cross-modality output $X_\mathrm{t \to b}$. 
% DTW. This is an algorithm for comparing the distances for each step of two time series data by brute force and finding a path with the shortest distance between time series, and the shortest path is the DTW distance.
Smaller the DTW-distances between $X_\mathrm{t}$ and $X_\mathrm{t \to b}$ means that the patterns are more similar and the Cross-VAE was able to replicate the original input time series. 
Thus, a smaller value is better. 
% The smaller the value, the closer the distance.
% , which means that the difference between the compared time series is small.

% \begin{table}[t]
% \begin{center}
% 	\caption{Quantitative evaluation of inter-conversion results. }
% 	\begin{tabular}{cccc} \hline
% 		& PSNR[dB] & SSIM & DTW \\ \hline 
%         Ours (LSTM) & 15.26 & 0.617 & 0.0411 \\
% 		Ours (Conv) & 15.99 & 0.707 & 0.0361 \\ 
% 		Average? & 9.197 & 0.159 & 0.206 \\ \hline
% 	\end{tabular}
% 	\label{tb:1}
% \end{center}
% \vspace{-2.5mm}
% \end{table}

\begin{table}[t]
\begin{center}
	\caption{Cross-Conversion Evaluations }
	\begin{tabular}{@{\extracolsep{4pt}}lccc@{}} 
	    \hline
		&\multicolumn{2}{c}{$Y_\mathrm{t \to b}$}& $Y_\mathrm{b \to t}$ \\
		\cline{2-3} \cline{4-4}
		& PSNR & SSIM & DTW \\ 
		\hline 
        Cross-VAE (LSTM) & 15.26 & 0.617 & 0.0411 \\
		Cross-VAE (Conv) & 15.99 & 0.707 & 0.0361 \\ 
		Class Average & 9.197 & 0.159 & 0.206 \\ 
        \hline
	\end{tabular}
	\label{tb:1}
\end{center}
\vspace{-4mm}
\end{table}

The results of quantitative evaluations are shown in Table \ref{tb:1}. 
In the table, we evaluate the difference between using LSTM layers and convolutional layers in the time series encoder and decoder. 
The results are compared to the images and time series of the average pattern in each respective class. 
% In the quantitative evaluation, the difference between the result of decoding the coded latent variable from the other test data and the test data was evaluated by each indicator.
% The fixed class intersection average to be compared was the average of the values calculated for each test data by using other test data which is the same existing class as the corresponding test data.
%
% The online $\to$ offline handwritten character conversion (inking) is quantified and the results are shown in table \ref{tb:1}.
%
% For both indices, the higher the value, the less the difference is in the compared images.
% Looking at the table, both Cross-VAE show better values than the results of comparing existing classes in the test data.
% In both PSNR and SSIM, Cross-VAE using Convolution was better than the same method using LSTM.
% The value of the fixed class intersection average is extremely small because it is thought that handwriting handwriting etc. have large blur even within the same class.
PSNR and SSIM are used for the cross-modal conversion from $X_\mathrm{t}$ to $Y_\mathrm{t \to b}$ and DTW is used for the evaluation of the cross-modal conversion from $X_\mathrm{b}$ to $Y_\mathrm{b \to t}$. 

For online to offline handwritten character conversion, or inking, the Cross-VAE did much better than the class average. 
In addition, the time series encoder and decoder with convolutional layers performed better than the LSTM. 
This shows that, despite being time series data, the convolutional layers were able to encode the information into the latent space better than the LSTM layers.

% Offline $\to$ Online handwritten character conversion (stroke order estimation) Quantitative evaluation is shown in the table.
% From the table, both Cross-VAEs showed better results than the class average.
% Also in the stroke order estimation, it was shown that conversion with Convolution to online handwritten character VAE is possible with high precision.
Similarly, for the offline to online handwritten character conversion, the Cross-VAE performed better than the average and the convolutional layer based time series encoder and decoder did better in reconstructing the time series. 
The DTW results specifically demonstrate that the Cross-VAE is able to predict the trajectories of the strokes. 
This information is normally lost during the rendering, however, the Cross-VAE is able to infer the stroke trajectory from the shared latent space. 

% From this weighing evaluation, it was found that in Cross-VAE, using Convolution for on-line handwritten character VAE shows good results both in inking and stroke order estimation.
% As mentioned above, LSTM is a model that has strengths in dealing with long-term time series dependence.
% Therefore, considering the task of this time, since it treats one character as one sample though it is using time series, the periodicity is not so much and the dependency on the past step in considering the next time step is not high.
% For this reason, we think that the model using Convolutilon has better results.

Both evaluations found that using convolutional layers was better than using LSTM layers. 
This is justified for this data target because handwritten characters are spatial coordinates where the relevance of every element depends on its neighbors. 
Structured data such as this is well suited to convolutional layers, whereas the advantages of maintaining long-term dependencies in LSTMs is lost. 
We believe that due to this, the convolutional layer based encoder and decoder for the time series modality produces better results.

\section{Conclusion}
In this paper, we proposed a VAE for mutual modality conversion called a Cross-VAE. 
The Cross-VAE is made from the merging of two VAEs of different modalities by enforcing a shared latent space.  
To train the Cross-VAE, we propose using the combination of reconstruction loss and distribution loss from the original VAE and an additional space sharing loss. 
The space sharing loss encourages the different modalities of the Cross-VAE to use the same latent space embedding.
% In this paper, we propose a network for mutual conversion of online / offline handwritten characters and conducted experiments.
% As a result, it was confirmed that mutual conversion between the time series data and the image data is possible by applying loss function considering mutual conversion to the network combining the two VAEs.
In the experiments, we used online and offline handwritten characters to verify the ability of the Cross-VAE. 
% In this way, online-to-online, online-to-offline, offline-to-offline, and offline-to-online conversion is done simultaneously.
The results show that the mutual conversion was possible and that the proposed Cross-VAE could accurately reconstruct the images and time series.

In the future, we will continue to improve the model and apply it to other applications. 
The Cross-VAE can be used for other types of data and tackle other tasks. 
Furthermore, this work opens the way for embedding different modalities into one shared latent space which can be used as a tool for representing those modalities in one space. 

% As future tasks, further improvement of the proposed network and consideration of application of this network can be mentioned.
% Specifically, we want to realize a character generation model with consideration of stroke characteristics by using this network as an evaluation function of a character generation model.

\section*{Acknowledgement}

This work was supported by JSPS KAKENHI Grant Number JP17H06100.

\newcommand{\BIBdecl}{\setlength{\itemsep}{0mm}}

\bibliographystyle{IEEEtran}
\bibliography{refer}

\begin{thebibliography}{10}
\providecommand{\url}[1]{#1}
\csname url@samestyle\endcsname
\providecommand{\newblock}{\relax}
\providecommand{\bibinfo}[2]{#2}
\providecommand{\BIBentrySTDinterwordspacing}{\spaceskip=0pt\relax}
\providecommand{\BIBentryALTinterwordstretchfactor}{4}
\providecommand{\BIBentryALTinterwordspacing}{\spaceskip=\fontdimen2\font plus
\BIBentryALTinterwordstretchfactor\fontdimen3\font minus
  \fontdimen4\font\relax}
\providecommand{\BIBforeignlanguage}[2]{{%
\expandafter\ifx\csname l@#1\endcsname\relax
\typeout{** WARNING: IEEEtran.bst: No hyphenation pattern has been}%
\typeout{** loaded for the language `#1'. Using the pattern for}%
\typeout{** the default language instead.}%
\else
\language=\csname l@#1\endcsname
\fi
#2}}
\providecommand{\BIBdecl}{\relax}
\BIBdecl

\bibitem{Tanaka_1999}
H.~Tanaka, K.~Nakajima, K.~Ishigaki, K.~Akiyama, and M.~Nakagawa, ``Hybrid
  pen-input character recognition system based on integration of online-offline
  recognition,'' in \emph{IAPR Int. Conf. Document Analysis and Recognition},
  1999, pp. 209--212.

\bibitem{Hamanaka}
M.~Hamanaka, K.~Yamada, and J.~Tsukumo, ``On-line japanese character
  recognition experiments by an off-line method based on
  normalization-cooperated feature extraction,'' in \emph{IAPR Int. Conf.
  Document Analysis and Recognition}, 1993, pp. 204--207.

\bibitem{Nguyen}
V.~Nguyen and M.~Blumenstein, ``Techniques for static handwriting trajectory
  recovery: A survey,'' in \emph{IAPR Int. Workshop on Document Analysis
  Systems}, 2010.

\bibitem{vae}
D.~P. Kingma and M.~Welling, ``Auto-encoding variational bayes,'' in \emph{Int.
  Conf. Learning Representations}, 2013.

\bibitem{gan}
I.~Goodfellow, J.~Pouget-Abadie, M.~Mirza, B.~Xu, D.~Warde-Farley, S.~Ozair,
  A.~Courville, and Y.~Bengio, ``Generative adversarial nets,'' in
  \emph{Advances in Neural Information Processing Systems}, 2014, pp.
  2672--2680.

\bibitem{baldi2012autoencoders}
P.~Baldi, ``Autoencoders, unsupervised learning, and deep architectures,'' in
  \emph{ICML Workshop Unsupervised and Transfer Learning}, 2012, pp. 37--49.

\bibitem{x-gacmn}
W.~Guo, J.~Liang, X.~Kong, L.~Song, and R.~He, ``X-gacmn: An x-shaped
  generative adversarial cross-modal network with hypersphere embedding,'' in
  \emph{Asian Conf. Computer Vision}, 2018.

\bibitem{Peng_2019}
Y.~Peng and J.~Qi, ``{CM}-{GANs},'' \emph{{ACM} Trans. Multimedia Computing,
  Communications, and Applications}, vol.~15, no.~1, pp. 1--24, feb 2019.

\bibitem{Spurr_2018}
A.~Spurr, J.~Song, S.~Park, and O.~Hilliges, ``Cross-modal deep variational
  hand pose estimation,'' in \emph{IEEE Conf. Computer Vision and Pattern
  Recognition}, 2018.

\bibitem{Huang_2018}
F.~Huang, X.~Zhang, C.~Li, Z.~Li, Y.~He, and Z.~Zhao, ``Multimodal network
  embedding via attention based multi-view variational autoencoder,'' in
  \emph{ACM Int. Conf. Multimedia Retrieval}, 2018.

\bibitem{serban2016multi}
I.~V. Serban, A.~G. Ororbia~II, J.~Pineau, and A.~Courville, ``Multi-modal
  variational encoder-decoders,'' in \emph{Int. Conf. Learning
  Representations}, 2016.

\bibitem{cyclegan}
J.-Y. Zhu, T.~Park, P.~Isola, and A.~A. Efros, ``Unpaired image-to-image
  translation using cycle-consistent adversarial networks,'' in \emph{{IEEE}
  Int. Conf. on Computer Vision}, 2017.

\bibitem{stargan}
Y.~Choi, M.~Choi, M.~Kim, J.-W. Ha, S.~Kim, and J.~Choo, ``{StarGAN}: Unified
  generative adversarial networks for multi-domain image-to-image
  translation,'' in \emph{{IEEE} Conf. on Computer Vision and Pattern
  Recognition}, 2018.

\bibitem{unit}
M.-Y. Liu, T.~Breuel, and J.~Kautz, ``Unsupervised image-to-image translation
  networks,'' in \emph{Advances in Neural Information Processing Systems},
  2017, pp. 700--708.

\bibitem{Nguyen_2010}
V.~Nguyen and M.~Blumenstein, ``Techniques for static handwriting trajectory
  recovery,'' in \emph{{IAPR} Int. Workshop Document Analysis Systems}.\hskip
  1em plus 0.5em minus 0.4em\relax {ACM} Press, 2010.

\bibitem{Kumar_Bhunia_2018}
A.~K. Bhunia, A.~Bhowmick, A.~K. Bhunia, A.~Konwer, P.~Banerjee, P.~P. Roy, and
  U.~Pal, ``Handwriting trajectory recovery using end-to-end deep
  encoder-decoder network,'' in \emph{Int. Conf. Pattern Recognition}, 2018.

\bibitem{Yu_Qiao_2006}
Y.~Qiao, M.~Nishiara, and M.~Yasuhara, ``A framework toward restoration of
  writing order from single-stroked handwriting image,'' \emph{{IEEE} Trans.
  Pattern Analysis and Machine Intelligence}, vol.~28, no.~11, pp. 1724--1737,
  nov 2006.

\bibitem{Zhao_2018}
B.~Zhao, M.~Yang, and J.~Tao, ``Pen tip motion prediction for handwriting
  drawing order recovery using deep neural network,'' in \emph{Int. Conf.
  Pattern Recognition}, 2018.

\bibitem{vrae}
O.~Fabius and J.~R. van Amersfoort, ``Variational recurrent auto-encoders,'' in
  \emph{ICLR Workshop}, 2014.

\bibitem{Akuzawa_2018}
K.~Akuzawa, Y.~Iwasawa, and Y.~Matsuo, ``Expressive speech synthesis via
  modeling expressions with variational autoencoder,'' in \emph{Interspeech},
  2018.

\bibitem{Bowman_2016}
S.~R. Bowman, L.~Vilnis, O.~Vinyals, A.~Dai, R.~Jozefowicz, and S.~Bengio,
  ``Generating sentences from a continuous space,'' in \emph{{SIGNLL} Conf.
  Computational Natural Language Learning}.\hskip 1em plus 0.5em minus
  0.4em\relax Association for Computational Linguistics, 2016.

\bibitem{pu2016variational}
Y.~Pu, Z.~Gan, R.~Henao, X.~Yuan, C.~Li, A.~Stevens, and L.~Carin,
  ``Variational autoencoder for deep learning of images, labels and captions,''
  in \emph{Advances in Neural Information Processing Systems}, 2016, pp.
  2352--2360.

\bibitem{ssvae}
D.~P. Kingma, S.~Mohamed, D.~J. Rezende, and M.~Welling, ``Semi-supervised
  learning with deep generative models,'' in \emph{Advances in Neural
  Information Processing Systems}, 2014, pp. 3581--3589.

\bibitem{wan2017variational}
Z.~Wan, Y.~Zhang, and H.~He, ``Variational autoencoder based synthetic data
  generation for imbalanced learning,'' in \emph{IEEE Symposium Series
  Computational Intelligence}, 2017, pp. 1--7.

\bibitem{unipen}
I.~Guyon, L.~Schomaker, R.~Plamondon, M.~Liberman, and S.~Janet, ``{UNIPEN}
  project of on-line data exchange and recognizer benchmarks,'' in \emph{Int.
  Conf. on Pattern Recognition}, vol.~2, 1994, pp. 29--33.

\bibitem{Noh_2015}
H.~Noh, S.~Hong, and B.~Han, ``Learning deconvolution network for semantic
  segmentation,'' in \emph{IEEE Int. Conf. Computer Vision}, 2015.

\bibitem{Hochreiter_1997}
S.~Hochreiter and J.~Schmidhuber, ``Long short-term memory,'' \emph{Neural
  Computation}, vol.~9, no.~8, pp. 1735--1780, 1997.

\bibitem{hinton2012neural}
G.~Hinton, N.~Srivastava, and K.~Swersky, ``Neural networks for machine
  learning lecture 6a overview of mini-batch gradient descent.''

\end{thebibliography}


% \end{thebibliography}

\end{document}